\documentclass[runningheads]{llncs}

\usepackage[T1]{fontenc}
\usepackage{graphicx}
\usepackage{amsmath}
\usepackage{booktabs}
\usepackage{xcolor}
\usepackage{changepage}
\usepackage{placeins}
\usepackage[hyphens]{url}
\usepackage[numbers]{natbib}

\title{Platonic Representations for Poverty Mapping:\\
Unified Vision-Language Codes \\ or Agent-Induced Novelty?}
\titlerunning{Platonic Representations for Poverty Mapping}
\author{Satiyabooshan Murugaboopathy\inst{1,2,3} \and
Connor T.\ Jerzak\inst{1,4} \and
Adel Daoud\inst{1,2,5}}
\authorrunning{S. Murugaboopathy et al.}
\institute{AI and Global Development Lab, \url{aidevlab.org} \and 
Institute for Analytical Sociology, Linköping University, Sweden \and
Institute of Computer Science, University Leipzig, Germany \and
Department of Government, The University of Texas at Austin, USA \and
NLP@DSAI, Chalmers University of Technology, Sweden
\email{satiyabooshan.murugaboopathy@liu.se, connor.jerzak@austin.utexas.edu, adel.daoud@liu.se}
}

\begin{document}

\maketitle
\thispagestyle{empty}
\pagestyle{empty}
\enlargethispage{4\baselineskip}
\begin{adjustwidth}{-0.2cm}{-0.2cm}
\begin{abstract}
We here investigate whether socio-economic indicators, such as household wealth, leave recoverable informational imprints in both satellite imagery (capturing physical features like buildings and roads) and Internet-sourced text (reflecting historical, cultural, and economic narratives of neighborhoods). Using Demographic and Health Survey (DHS) data from African neighborhoods (clusters), we pair high-resolution Landsat images with textual descriptions generated by large language models (LLMs) conditioned on location and year, as well as text retrieved by an LLM-driven AI Search Agent from web sources.
We develop a multimodal framework that predicts household wealth (measured by the International Wealth Index (IWI)) through five pipelines: (i) a vision model on satellite images, (ii) an LLM using only location and year, (iii) an AI agent that searches and synthesizes web text, (iv) a joint image-text encoder, and (v) an ensemble of all signals.
Our framework yields three contributions. {\it First,} evaluations show that fusing vision and agent/LLM-generated text improves on vision-only baselines in wealth prediction (e.g., $R^2=0.77$ vs. $0.63$ on out-of-sample splits), with LLM-internal knowledge (artificial neural memory) proving surprisingly predictive in out-of-country and out-of-time generalization. 
{\it Second,} we find suggestive evidence of partial representational alignment: fused embeddings from vision and language modalities correlate moderately (median cosine similarity across modalities of about 0.60 after alignment). This pattern is broadly consistent with the Platonic Representation Hypothesis, but it does not by itself establish convergence to a single shared latent representation. Because agent-retrieved data yields only marginal and unstable gains across splits, our evidence for the Agent-Induced Novelty Hypothesis is limited.
{\it Third,} we release a large-scale multimodal dataset comprising approximately 60,000 DHS clusters, each linked to satellite images, LLM-generated descriptions, and associated texts retrieved by AI agents.
\par\smallskip
\noindent\keywordname\enspace Poverty Mapping; Satellite Imagery; Large Language Models; Multimodal Learning; AI Agents; Earth observation; Remote sensing\par
\par\smallskip
\noindent{\footnotesize\textbf{Citation:} Satiyabooshan Murugaboopathy, Connor T.~Jerzak, and Adel Daoud (2026). \emph{Platonic Representations for Poverty Mapping: Unified Vision-Language Codes or Agent-Induced Novelty?} In \emph{Proceedings of the 7th International Conference on Social Computing (ICSC 2026)}, Communications in Computer and Information Science. Springer Nature.\par}
\end{abstract}
\end{adjustwidth}

\section{Introduction}\label{s:Intro}

In an era where the timely and accurate measurement of socio-economic disparities is crucial for effective policy-making, humanitarian aid, and sustainable development, traditional household surveys, such as the Demographic and Health Surveys (DHS), provide invaluable ground-truth data but face significant limitations in terms of scale, cost, and frequency \citep{dhs2013demographic}. These surveys, conducted periodically across low- and middle-income countries, capture wealth metrics such as the International Wealth Index (IWI)---a composite measure of household assets and living conditions---but often leave significant gaps in geographic and temporal coverage, particularly in remote or rapidly changing regions \citep{sakamoto2025scopingreviewearthobservation}. To bridge these gaps, researchers have increasingly turned to remote sensing technologies, leveraging high-resolution Earth observation (EO) imagery from satellites, such as Landsat, to infer poverty patterns through visual cues, including infrastructure density, land use, and vegetation health \citep{yeh2020using,pettersson2023time,kakooei2024mappingafricasettlementshigh,burkeUsingSatelliteImagery2021}. While these vision-based approaches have shown promise, they are inherently limited by what can be ``seen'' from above, often missing nuanced socio-cultural, historical, or contextual factors that influence material well-being \citep{o2023seeing,pmlr-v275-zhu25a}. 

The advent of large language models (LLMs) and multimodal AI systems offers a complementary pathway, enabling the extraction and synthesis of textual information from vast digital repositories, including web sources and encyclopedic knowledge \citep{sarmadi2025leveragingchatgptsmultimodalvision}. This raises two interrelated questions: (1) To what extent can useful textual information about neighborhoods in low- and middle-income countries be recovered from LLMs' artificial neural memory or found by AI Search Agents on the Internet? (2) If recoverable, how well can socio-economic status be distilled into a compact, latent representation across vision and language modalities, potentially converging toward a shared ``Platonic'' representation of material well-being \citep{huh2024position}? Or, do modalities complement each other to enhance estimation?

\begin{figure}[h]
\centering
\includegraphics[width=0.6\linewidth]{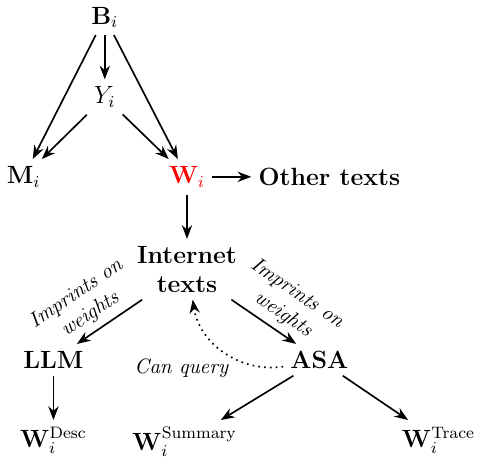}
\caption{
DAG representing the imprint of the poverty/wealth index \(Y_i\) on satellite imagery \(\mathbf{M}_i\) and textual data \(\mathbf{W}_i\), processed by LLM's neural weights and queried by AI Search Agent. \(\mathbf{W}_i\) is challenging to observe directly, hence colored in red. Instead, the goal is to reconstruct a faithful representation via the LLM's neural memory or AI agent search capabilities. These texts are denoted $\mathbf{W}_i^{\text{Desc}}$ for the LLM, and $\mathbf{W}_i^{\text{Trace}}$ (raw search result) and $\mathbf{W}_i^{\text{Summary}}$ (summary of search result) for the search agent. \(\mathbf{B}_i\) represents background factors.}\label{fig:SimpleDAG}
\end{figure}

Figure~\ref{fig:SimpleDAG} illustrates these questions via a directed acyclic graph (DAG), where the materialization of poverty/wealth index \(Y_i\) in African neighborhoods causally influences both satellite imagery \(\mathbf{M}_i\) and textual data \(\mathbf{W}_i\). Socio-economic conditions leave observable traces in the physical environment (e.g., building density, road networks) visible in EO data, as well as in linguistic artifacts (e.g., historical accounts, news reports) \citep{daoud2019international,daoudStatisticalModelingThree2023}. However, \(\mathbf{W}_i\) is challenging to observe directly, necessitating LLMs to reconstruct it from neural memory as \(\mathbf{W}_i^{\text{Desc}}\), and AI Search Agents (ASA) to query the Internet for \(\mathbf{W}_i^{\text{Trace}}\) (raw traces) and thereafter generate \(\mathbf{W}_i^{\text{Summary}}\) (summaries).

Despite growing interest in multimodality in poverty mapping, systematic investigations remain scarce, hampered by the lack of aligned, large-scale datasets \citep{lamichhane2025exploring,kakooei2024analyzingpovertyintraannualtimeseries}.
To address this limitation, we here analyze whether household wealth on the African continent, as measured by DHS IWI scores, can be textually and visually reconstructed and encoded in a joint latent space. Drawing on a continent-scale corpus of more than 60,000 DHS clusters across Africa from 1990 to 2020, we pair high-resolution, cloud-free Landsat composites with LLM-generated spatiotemporal narratives and AI-agent-retrieved contextual information. At its heart, our approach taps five different signals to ``see'' and describe each neighborhood’s wealth. First, we use satellite images to capture physical clues---things like roads, buildings, and vegetation. Next, we ask a language model to imagine a narrative for the place based only on its location and year. Then, an AI agent goes online, gathers real-world text about the area, and boils it down into a concise summary. After that, we train a joint encoder to blend the visual and textual cues into a single shared representation. Finally, we let an ensemble weigh and combine all five perspectives before we analyze embedding spaces to investigate representational convergence and complementarity across modalities. Figure~\ref{fig:PipeViz} provides a visual overview.

The remainder of the paper is organized as follows: We first review the related work on remote sensing and AI in poverty mapping. Next, we detail our agent and data curation frameworks, followed by an experimental performance and representational analysis, and conclude with implications, limitations, and future directions. 

\section{Problem Setup \& Related Work}\label{s:Related}

Early efforts in remote sensing for poverty estimation used nighttime light imagery as a proxy for economic activity \citep{elvidge2009global}. More recent advances have focused on daytime satellite imagery combined with deep learning techniques. For example, Jean et al.~\citep{jean2016combining} demonstrated the use of transfer learning from night lights to predict poverty from daytime images in African countries. Subsequent works have applied convolutional neural networks directly to satellite data for wealth prediction \citep{yeh2020using,pettersson2023time,kakooei2024analyzingpovertyintraannualtimeseries}. Interpretability has also been a focus, with methods using object detection to generate explainable poverty maps \citep{babenko2017povertymappingusingconvolutional}. Recent reviews synthesize the state of Earth observation and ML for poverty research, highlighting applications in causal inference and small-area estimation \citep{sakamoto2025scopingreviewearthobservation}. Other studies explore fairness and biases in satellite-based poverty maps \citep{aiken2023fairness}. 

The emergence of large language models (LLMs) has opened new avenues for socio-economic inference using textual data. LLMs have been employed to estimate regional socio-economic indicators directly from prompts \citep{han2024geosee}. Synergizing LLM agents with knowledge graphs has shown promise for socioeconomic prediction \citep{zhou2024synergizingllmagentsknowledge}. Additionally, biases in LLMs related to socioeconomic attributes have been investigated \citep{arzaghi2024understandingintrinsicsocioeconomicbiases,depieuchon2025benchmarkingdebiasingmethodsllmbased}.

Along with LLMs, multimodal approaches that combine vision and language are increasingly being explored. Sarmadi et al.~\citep{sarmadi2025leveragingchatgptsmultimodalvision} leveraged GPT 4's multimodal capabilities to rank satellite images by poverty levels. Other works combine satellite imagery with non-visual features (such as X/Twitter activity, distance from residential roads, and Internet speed) to improve poverty prediction \citep{JungLastMile}. Despite these advances, systematic studies on representational convergence across modalities in poverty mapping remain limited \citep{lamichhane2025exploring}.

Our work builds on the framework outlined in the Introduction (Figure~\ref{fig:SimpleDAG}), where the latent poverty/wealth index \(Y_i\) causally influences both satellite imagery \(\mathbf{M}_i\) and textual data \(\mathbf{W}_i\), assuming conditional independence between modalities given \(Y_i\). This structure motivates our investigation into whether vision and language encoders converge toward a shared ``Platonic'' representation of material well-being, as inspired by the Platonic Representation Hypothesis proposed by Huh et al.~\citep{huh2024position}. While this hypothesis has been examined in general vision-language models \citep{radford2021learning}, its application to specific scientific domains, such as economic development and poverty mapping---particularly in contexts where data modalities are gathered endogenously by AI agents---is novel. 

\paragraph{Agent-Induced Novelty Hypothesis.}
We also examine whether dynamic, agent-driven data collection can introduce predictive information beyond that captured by static LLM memory. In agentic systems, AI agents autonomously gather and synthesize data via LLM-guided paths. We refer to the empirical possibility that agent-gathered data adds complementary structure beyond LLM-only descriptions as the {\it Agent-Induced Novelty Hypothesis}. This hypothesis does not require agent text to form a wholly separate latent representation; it only predicts that agent retrieval should add stable predictive signal after conditioning on vision and LLM-only text.

Unlike static or LLM-internal representations, agent-induced data evolves through iterative queries, retrievals, and syntheses \citep{miehling2025agentic}. This process could introduce useful complementarities, but it could also add retrieval noise, duplicated background knowledge, or post-processing artifacts. We therefore test the novelty hypothesis by comparing embeddings from agent traces (\(\mathbf{W}_i^{\text{Trace}}\)) against LLM-only descriptions (\(\mathbf{W}_i^{\text{Desc}}\)) and by asking whether fused models gain stable performance from agent traces beyond the vision and NMR baselines.

\paragraph{Formalization.}

To test the Platonic Representation Hypothesis and the Agent-Induced Novelty Hypothesis outlined above, we formalize the poverty mapping task as a supervised regression problem. The objective is to predict the International Wealth Index \(Y_i \in [0, 100]\) for each DHS cluster \(i\), which is aggregated at the neighborhood level from household surveys. Let \(\mathbf{M}_i\) denote the Earth observation (EO) features extracted from satellite imagery (e.g., Landsat multispectral bands, processed via a pre-trained vision model to yield embeddings).

Textual signals, as introduced in the DAG (Figure~\ref{fig:SimpleDAG}) and elaborated in the hypotheses, are decomposed into: \(\mathbf{W}_i^{\text{Trace}}\), the raw agent search traces (e.g., concatenated Wikipedia excerpts and web search results); \(\mathbf{W}_i^{\text{Summary}}\), the agent's synthesized summary of those traces; \(\mathbf{W}_i^{\text{Desc}}\), the LLM-only spatiotemporal description (e.g., generated from location-year prompts without external search); and scalar predictions, \(\hat{Y}_i^{\text{Agent}}\) and \(\hat{Y}_i^{\text{LLM}}\), from agent and LLM-only decoding, respectively. Embeddings from textual components (e.g., via a language model encoder) are denoted \(\mathbf{E}_i^{\text{Text}}\), while fused multimodal embeddings are \(\mathbf{E}_i^{\text{Fused}} = f(\mathbf{M}_i, \mathbf{E}_i^{\text{Text}})\) for some encoding function \(f\).

\section{Data \& Methods}\label{s:DataMethods}

\paragraph{Data Curation.}
We create a multimodal dataset centered on DHS units, comprising approximately 60,000 geolocated neighborhood clusters across Africa, with surveys conducted between 1990 and 2020. 

For each cluster, we extract the International Wealth Index (IWI) as ground truth, a composite score (0--100) reflecting household assets and conditions, which is aggregated to the cluster (neighborhood) level. 

Visual data consist of high-resolution Landsat composites (30m resolution) centered on DHS coordinates. We extract 224$\times$224 pixel tiles using  red, green, and blue 
%shortwave infrared, and longwave infrared
channels. 
Following conventional Landsat preprocessing, we mask clouds and cloud shadows using quality-assurance bands and form median temporal composites centered on the survey year.

Text data are generated via two channels: (i) LLM-generated descriptions (GPT-4.1 Nano, no search or tools enabled) conditioned on year,  location coordinates, and location name (reverse-geocoded from coordinates); and (ii) AI-agent-retrieved context, where an LLM-driven agent queries Wikipedia and Internet search to extract socioeconomic, historical, and contextual information about DHS place name, also given year and coordinates. 

The AI Search Agent has a GPT-4.1 Nano LLM core, and was constructed using the open-source tool LangGraph \citep{wang2024agent}; the maximum recursion depth was set to 20, meaning that the agent could take at most 20 steps of iterative searching before completing. In practice, we observed that the majority of agent invocations terminated with three search steps or fewer. (Note that both the LLM and the search agent use the same backbone and identical place data, isolating the impact of agentic capabilities.)

The resulting combined corpus---images, LLM texts, agent traces, with IWI labels, dubbed \texttt{IWI-Africa-Multimodal}---will be released on Hugging Face.\footnote{To link with the full DHS data, users must register with the DHS Program and agree to its privacy and data use policies, as outlined at \texttt{https://dhsprogram.com}.} See Table \ref{tab:agent-trace-examples} for two illustrative AI Search Agent traces. Table \ref{tab:domain-counts} displays the most commonly found URLs in the AI search. 

\begin{table}[h]
\centering
\caption{
Examples of data gathered by the AI Search Agent.
}
\label{tab:agent-trace-examples}
\tiny
%\hspace{-1.99cm}
\begin{tabular}{
p{0.75cm} p{0.75cm} p{1.55cm} p{2.75cm} p{2cm} p{2.25cm}}
\hline
\textbf{Lat} & \textbf{Long} & \textbf{Place Name} & \textbf{Full Agent Trace} & \textbf{Wikipedia Trace} & \textbf{Search Trace 1/10} \\
\hline
-18.85 & 47.58 & Manazary, Madagascar & \texttt{\tiny Manazary is a small rural commune in the outskirts of Antananarivo... suggesting a moderate level of development typical of rural Madagascar. [concatenated with search results: Antananarivo- Avaradrano is a district... elevation of 1,318 metres... etc.]} & \texttt{\tiny Manazary is a commune in Madagascar... population estimated at 37,000 in 2001... primarily engaged in agriculture with rice as main crop... 40\% in fishing.} & \texttt{\tiny Antananarivo- Avaradrano is a district of Analamanga in Madagascar. It covers smaller communes in the outskirts of Antananarivo...}  \\
\hline
-1.63 & 29.36 & Kanzenze, Rwanda & \texttt{\tiny Rubavu District, including Kanzenze sector, experienced social and economic development... suggests a medium level of wealth and infrastructure. [concatenated with search results: Gender equality on socio-economic development in Rubavu... population and housing census... etc.]} & \texttt{\tiny Rubavu District is one of seven districts in Western Province, Rwanda... capital is Gisenyi... urban area population of 149,209 in 2012.} & \texttt{\tiny APPROVAL SHEET This thesis entitled "Gender Equality on socio-economic development in Rubavu district, Rwanda"... } \\
\hline
\end{tabular}
\end{table}

\begin{table}[tb]
\caption{Distribution of primary external text sources retrieved by the AI Search Agent across the full \texttt{IWI-Africa-Multimodal} dataset. Wikipedia dominates the retrieved traces, reflecting its role as the most comprehensive open-knowledge repository for African localities.}
\label{tab:domain-counts}
\centering
\small
\begin{tabular}{l r}
\hline\hline\toprule
\textbf{Domain}                  & \textbf{Cluster Count} \\
\hline
\midrule
wikipedia.org                    & 66,998 \\
mapcarta.com                     & 28,337 \\
researchgate.net                 & 26,375 \\
citypopulation.de                & 18,705 \\
academia.edu                     & 16,333 \\
city-facts.com                   & 12,395 \\
opendataforafrica.org            & 12,062 \\
worldbank.org                    & 10,983 \\
\bottomrule
\hline
\end{tabular}
\end{table}

\begin{figure}
    \centering
    \includegraphics[width=\linewidth]{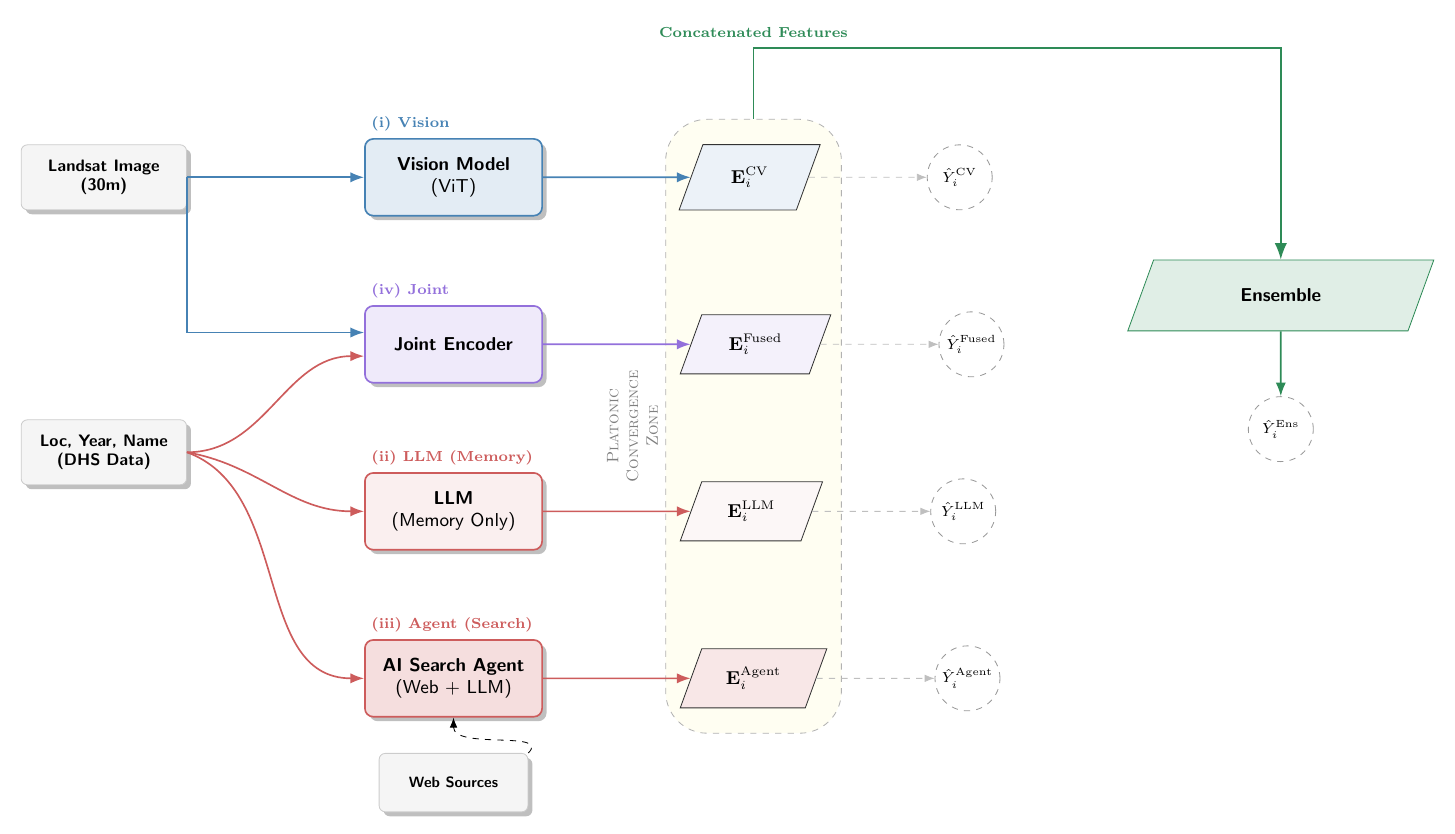}
    \caption{
    Overview of the quintuple prediction framework for estimating household wealth (IWI) from DHS clusters.
    Vision, LLM-memory, agent-search, joint-encoding, and ensemble pipelines produce complementary candidate predictors.
    }\label{fig:PipeViz}
\end{figure}

\paragraph{Prediction Framework.}
Figure \ref{fig:PipeViz} illustrates our implementation of the five pipelines for estimating IWI scores while evaluating convergence and complementarity across modalities.

(i) \textsc{Vision Pipeline}: A Vision model (e.g., a 12-layer Vision Transformer architecture with patch size 16~\citep{stewart2022torchgeo}) is pre-trained unsupervised on Landsat images, then used to encode the 224$\times$224 five-channel Landsat composites. We predict IWI from the resulting visual embeddings with ridge regression.

(ii) \textsc{LLM-Only Pipeline}: An LLM (e.g., Llama-4-Maverick, GPT-4.1 Nano) predicts IWI directly from location-year prompts, leveraging internal neural memory; outputs include prediction, justification of prediction, and confidence in prediction.

(iii) \textsc{AI Search Agent Pipeline}: An AI agent (GPT-4.1 Nano core) retrieves and synthesizes web text (as above), then predicts IWI; we extract traces (raw text) and justifications for embedding.

(iv) \textsc{Ensemble Pipeline}: Pipelines (i)–(iii) independently generate modality-specific embeddings, which are then concatenated and used to train a ridge regression model supervised on IWI labels.

Here, embeddings refer to dense vector representations of input data (e.g., images or text) learned by neural networks, which capture semantic or visual features in a compact, latent space suitable for downstream tasks like regression. In our pipelines, we distinguish between (1) frozen embeddings, which are pre-computed from pre-trained models---e.g., OpenAI's \texttt{text-embedding-3-small} [abbreviated as OAI] or \texttt{sentence-transformers/all-mpnet-base-v2}, [abbreviated as MPNet]---and not updated during training to preserve general knowledge, and (2) fine-tuned embeddings, which are adjusted on our dataset to adapt to poverty-specific patterns. This distinction allows us to evaluate the trade-offs between leveraging off-the-shelf representations and task-specific optimization.

We also define modality-specific acronyms used across experiments: NMR (Neural Memory Reconstruction) denotes the LLM-only pipeline, where text is generated solely from the model's internal knowledge (artificial neural memory) based on location and year prompts; ASA (AI Search Agent) refers to the pipeline using agent-retrieved Internet text, including raw traces and summaries; and CV (computer vision) indicates the satellite imagery pipeline. 

We conduct three distinct evaluation experiments designed to assess model robustness across spatial and temporal dimensions:
\begin{itemize}
\item \textsc{Random Split:} A standard 80/20 train-test split is applied without any geographic or temporal constraints, serving as the baseline for performance comparison.
\item \textsc{Out-of-Country (OOC):} The model is trained on clusters from a subset of countries (a random 80\% subset) and evaluated on held-out countries not present in the training set (the remaining 20\%). Countries, rather than individual clusters, are the sampling units. This test evaluates cross-border generalization and the model's ability to transfer knowledge beyond national boundaries.
\item \textsc{Out-of-Time (OOT):} The model is trained on data from one time period (encompassing 80\% of the data) and evaluated on a disjoint time span (the remaining 20\%). Complete survey years, rather than individual clusters, are the sampling units. This assesses temporal generalization and the model's resilience to shifts in socio-economic conditions over time.
\end{itemize}    

For each experiment, we ensure strict data partitioning to prevent information leakage. In the OOC split, entire countries are treated as atomic units and assigned to either training or test folds. We randomly assign countries such that the distribution of clusters across folds is balanced. In the OOT split, complete years are used as units of partitioning, ensuring that no overlapping time periods exist between the train and test sets. In the \textsc{Random Split}, cluster assignments are made purely at random, with no restrictions based on location or year. For all repeated splits, we inspect train/test balance in mean IWI, cluster counts, country size, and survey-year coverage.

All models are evaluated using two distinct training strategies, depending on whether embeddings are frozen or fine-tuned:
\begin{itemize}
\item For models with \textit{frozen embeddings}, we perform 100 bootstrap iterations with an 80/20 train-test split per bootstrap iteration. Final results are reported as the mean and standard error across bootstraps.
    
\item For models with \textit{fine-tuned embeddings}, we use 5-fold cross-validation, with a 70/15/15 train/validation/test split per fold. This ensures stable and reliable performance estimates while minimizing overfitting. Uncertainties are estimated as the standard deviation of the performance metrics obtained across the five cross-validation iterations on the test fold.
\end{itemize}

Each model is evaluated under the same split protocol, allowing for comparisons between single-modality and multimodal configurations. All evaluations are conducted using single-frame inputs; no temporal image or text sequences are used. This design choice allows us to isolate the impact of modality fusion from temporal dynamics.

Performance is measured using the coefficient of determination ($R^2$) and root mean squared error (RMSE) on predicting out-of-sample poverty. The $R^2$ metric quantifies the proportion of variance in the International Wealth Index (IWI) that is explained by the model, providing a clear interpretation of the model's predictive power. Confidence intervals are obtained via bootstrapping (for frozen embedding approaches) or cross-validation (for fine-tuned embedding approaches).

This experimental setup enables an assessment of how multimodal signals---particularly those generated through agent-driven web retrieval and LLM-based reasoning---enhance poverty prediction beyond what is achievable with satellite imagery alone. It also supports our investigation into the representational convergence of vision and language, testing both the Platonic Representation Hypothesis and the Agent-Induced Novelty Hypothesis in a real-world, open-ended context.

\section{Results}\label{s:Results}

\begin{figure}[!htbp]
  \centering
  \includegraphics[width=0.75\linewidth]{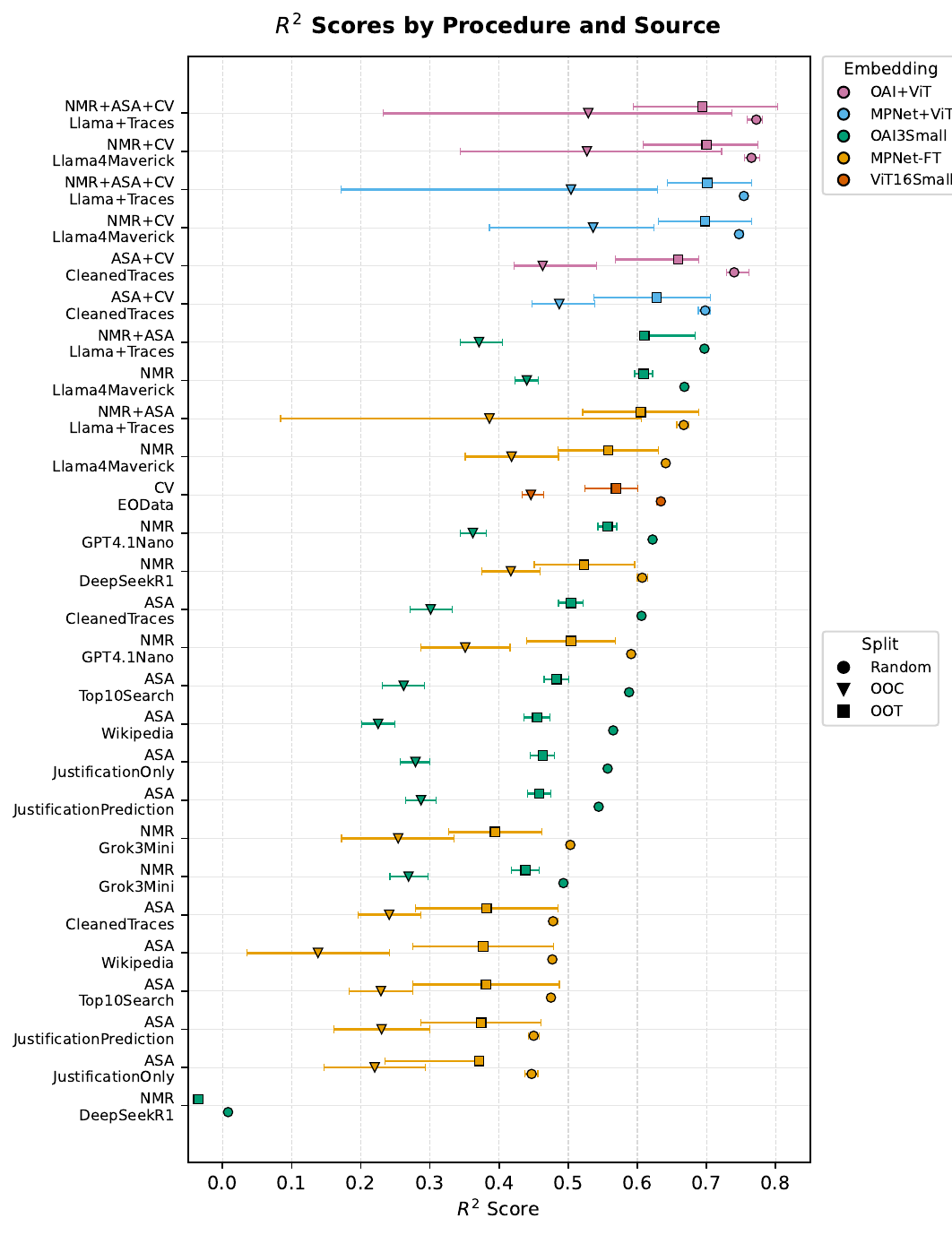}
  \caption{
    Test performance ($R^2$) across evaluation splits (\textsc{Random}, OOC, OOT), grouped by model procedure and data source.
    Each dot represents the $R^2$ for a model under a specific split strategy.
    The dot shape encodes the split type, and the dot color encodes the embedding model.
    Rows are ordered by highest $R^2$ for readability.
    This visualization highlights the performance of various model+embedding combinations across generalization scenarios.
  }
  \label{fig:test_split_scatter}
\end{figure}

\paragraph{General patterns} Our evaluation reveals performance gains when combining multi-modal signals for poverty prediction, as shown in Figure~\ref{fig:test_split_scatter} (full results in Table \ref{tab:test_split_sorted_by_modality}, with uncertainty estimates in Table \ref{tab:ci95_bounds_summary}). The NMR+CV approach using Llama-4-Maverick with OpenAI embeddings achieves the highest performance across all evaluation strategies, with an $R^2$ of 0.765 on the random split. This improves on the best single-modality approach (NMR alone at $R^2$ = 0.668) and on the CV-only baseline ($R^2$ = 0.634). The CV-only baseline also compares favorably with the shallow baseline reported by Pettersson et al.~\citep{pettersson2023time} ($R^2$ = 0.60 on their OOC split), but that comparison should be interpreted cautiously because the datasets, label definitions, split construction, and image inputs differ; our CV model uses single-frame daytime imagery, whereas Pettersson et al.~\citep{pettersson2023time} study time-series imagery and may include both daytime and nighttime inputs. The performance gains are consistent across evaluation strategies, with the combined model maintaining strong performance even under OOC and out-of-time (OOT) splits. It is notable that the ASA searchers alone perform below the CV-only benchmark, possibly indicating that finding relevant online information is a noisy process.

Notably, the OOC split (where no countries in the training set appear in the test set) produces the most significant performance degradation across all models, indicating that country-specific features play a critical role in poverty prediction. This is particularly evident in the CV-only baseline, which drops from $R^2$ = 0.634 (\textsc{Random}) to $R^2$ = 0.446 (OOC). In contrast, the OOT split (where no years in the training set appear in the test set) shows a relatively smaller performance drop, suggesting that year-specific features are less critical than country-specific features for poverty prediction in our dataset.

\FloatBarrier

\begin{figure}[!htbp]
  \centering
  \includegraphics[width=0.95\linewidth]{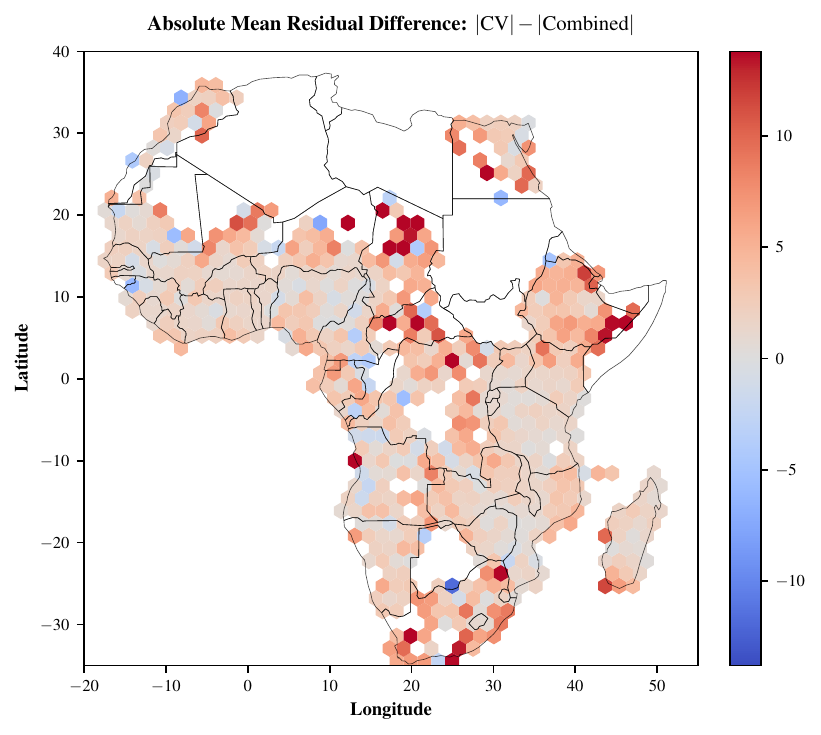}
  \caption{
    Spatial distribution of the difference in prediction residuals between CV-only and NMR+CV.
    \textbf{\color{blue} Blue} indicates regions where the baseline CV model outperforms NMR+CV. \textbf{\color{red}Red} indicates regions where NMR+CV outperforms CV.
    Hexagons summarize values across locations.
  }
  \label{fig:residual_map}
\end{figure}

To further contextualize performance, we conducted an extensive analysis of different model architectures and data sources, building on the textual representations defined earlier: $\mathbf{W}_i^{\text{Trace}}$ (raw agent search traces), $\mathbf{W}_i^{\text{Summary}}$ (agent-synthesized summaries), and $\mathbf{W}_i^{\text{Desc}}$ (LLM-generated descriptions). Here, \texttt{CleanedTraces} refers to all agent-crawled text data ($\mathbf{W}_i^{\text{Trace}}$) without filtering; \texttt{Wikipedia} is the subset from Wikipedia sources; \texttt{JustificationOnly} includes only the agent's justification of IWI prediction text; and \texttt{JustificationPrediction} combines justification with the agent's scalar prediction. The results reveal that the \texttt{CleanedTraces} approach consistently outperforms all other subset text sources. This suggests that the full breadth of agent-generated context provides richer signals for poverty prediction.

The ASA+CV approach using \texttt{CleanedTraces} achieves an $R^2$ of 0.740 on the random split, which is higher than the CV-only baseline ($R^2$ = 0.634) and the best single-modality NMR approach ($R^2$ = 0.668). This suggests that the AI-search agent's raw text collection can complement the vision pipeline, although the gain does not by itself establish that agent retrieval introduces a stable new representation. See Table \ref{tab:test_split_sorted_by_modality} for full results.

Our analysis of different embedding models reveals that \texttt{text-embedding-3-small} (OAI) consistently outperforms the fine-tuned MPNet model (\texttt{all-mpnet-base-v2}) across all evaluation strategies (see Table~\ref{tab:test_split_sorted_by_modality} and Figure~\ref{fig:test_split_scatter} for detailed results). For instance, in the NMR+CV pipeline under random splits, OAI achieves an $R^2$ of 0.765 compared to MPNet's 0.747. MPNet cannot match the performance of the frozen-weights OAI model. This suggests that the pretraining and contextual understanding captured in OAI provide a significant advantage for poverty prediction tasks over a smaller model fine-tuned on our dataset.

\paragraph{Africa-wide Spatial and Temporal Analysis of Model Performance.}
To quantify spatial variability in gains from multimodality, Figure~\ref{fig:residual_map} plots  the following difference in absolute residuals, $(\lvert Y_i-\hat{Y}_i^{\text{CV}} \rvert - \lvert Y_i-\hat{Y_i}^{\textrm{Best} }\rvert)$. This value is negative (\textbf{\color{blue}blue}) when the CV baseline outperforms the best multi-modal approach; it is positive (\textbf{\color{red}red}) otherwise.
We benchmark here the best-performing joint model against the computer-vision model, as the CV approach is dominant in the ML literature on poverty prediction.

As visualized in Figure~\ref{fig:residual_map}, the combined NMR+CV model demonstrates particularly strong performance in densely populated regions such as South Africa, where it achieves improved accuracy compared to the CV-only baseline. It also significantly outperforms in conflict-affected areas of Somalia and central Africa (e.g., Chad). In contrast, performance improvements are more modest along the eastern coast (e.g., Guinea), suggesting that the information provided by the NMR component may be less valuable in regions with moderate levels of development and inequality.

\FloatBarrier

\begin{figure}[!htbp]
  \centering
  \includegraphics[width=0.78\linewidth]{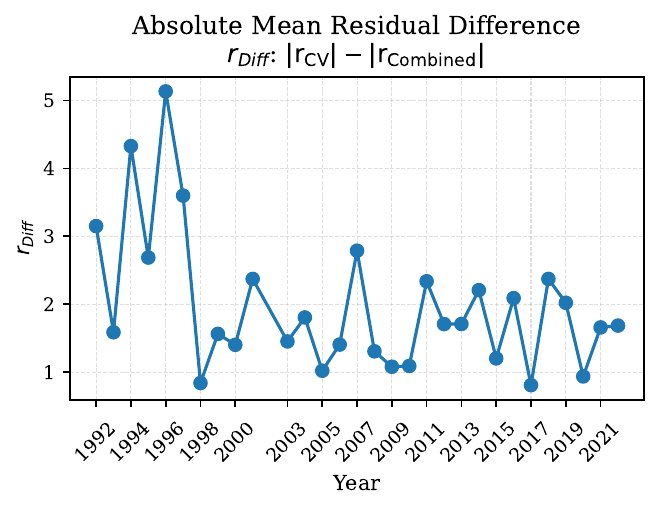}
  \caption{
    Absolute mean residual difference across years for the best system: NMR+ASA+CV vs. CV-only baseline.
    }
  \label{fig:residuals_years_cv_comb}
\end{figure}

\paragraph{Time-series improvements.} To complement this spatial analysis, we also analyzed where the temporal improvements are occurring, using the same metric as for the Africa-wide spatial evaluation. Figure \ref{fig:residuals_years_cv_comb} shows that the improvements are consistent over time, where the multi-modal model outperforms the CV-only model. Most residual improvements occurred in the 1990s. Interestingly, this is the period when satellite images are most scarce, with an average pixel availability of about three-quarters across the continent, from 1990 to 2000 \cite{jerzak2023integrating}. 

\FloatBarrier

\paragraph{Model Size Analysis.}

The performance of different LLMs reveals a clear relationship between model size and performance, with the largest model (Llama-4-Maverick, 405B parameters) achieving the best results. However, the smaller models (GPT-4.1 Nano, Grok-3-Mini) perform comparably well at significantly lower computational cost. This suggests that for practical deployment, moderately sized LLMs can provide an excellent balance between performance and cost.

\paragraph{Agent-Induced Novelty vs. Platonic Representation.}
One of the most notable findings from our experiments is that the LLM-only NMR approach consistently outperforms the ASA approaches. For example, NMR with Llama-4-Maverick achieves up to $R^2$ = 0.668, while the best ASA approach (\texttt{CleanedTraces}) achieves $R^2$ = 0.606. This pattern suggests that LLM-internal knowledge is highly predictive, but the comparison is not a controlled test of internal knowledge versus retrieval. The NMR and ASA pipelines differ in model core, input length, retrieval process, text noise, and post-processing. 
The result is interpreted cautiously; further evidence would be required to establish a general conclusion that neural memory is always more predictive than agent-retrieved text.
That said, our findings are also consistent with findings by Gema et al.~\citep{gema2025inverse}, which show reduced performance in language tasks as context (e.g., via reasoning chains) becomes overly large.

The marginal gain from NMR+ASA+CV ($R^2 = 0.772$) over NMR+CV ($R^2 = 0.765$) in random splits is small and unstable across evaluation settings (e.g., absent in OOT splits). We therefore treat the evidence for the Agent-Induced Novelty Hypothesis as limited. The results are more consistent with partial cross-modal alignment: language and vision signals appear to encode overlapping information about material well-being, while agent data contributes at most modest complementary signal in specific scenarios. Future work should isolate agent-specific contributions through more controlled retrieval ablations, matched input lengths, and embedding-distance tests.

\paragraph{Validation of Intra-Cluster Similarity Against Null Distribution.}
\begin{figure}[!htbp]
  \centering
  \includegraphics[width=0.60\linewidth]{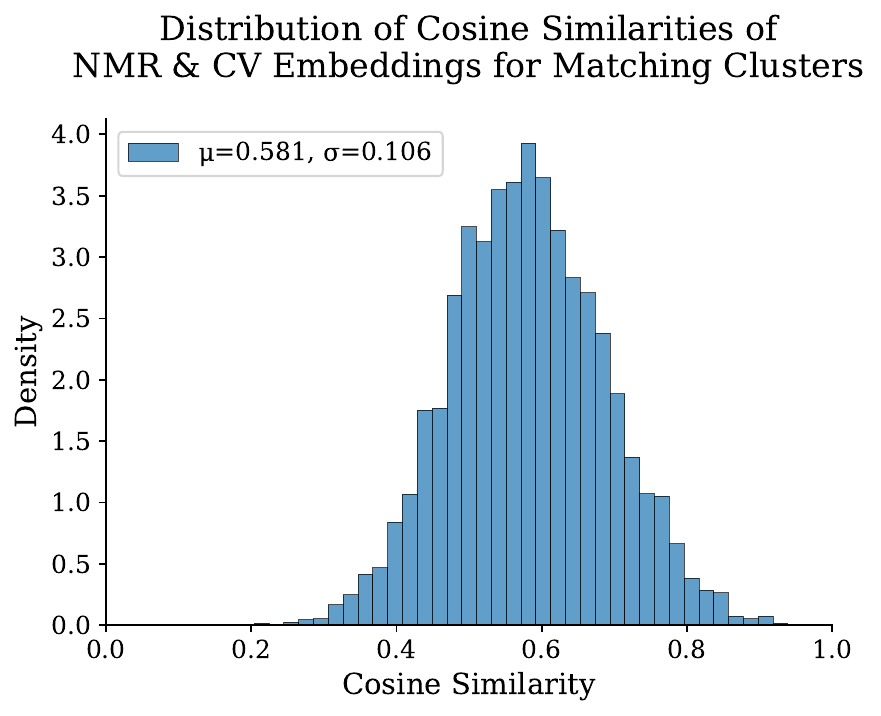}
  \caption{
    Histogram of cosine similarities between NMR and CV embeddings, after alignment through canonical correlation (first component), for matched DHS clusters. 
  }
  \label{fig:hist_cos_sim}
\end{figure}

To statistically validate the observed alignment in Figure~\ref{fig:hist_cos_sim}, we fit the canonical correlation analysis (CCA) alignment on the training split and evaluate cosine similarities on held-out clusters. We compare matched NMR--CV pairs against a null distribution constructed from random pairings, including shuffled location pairs within the same country and survey year when such matches are available. A one-sample $t$-test against the random-pairing baseline produced a $t$-statistic of 312.96 and a $p$-value $< 1\mathrm{e}{-10}$. This evidence suggests partial representational alignment across modalities, but it should not be interpreted as proof that vision and text converge to a single shared latent representation.

Our median cosine similarity of approximately 0.60 between Earth Observation (EO) and LLM embeddings indicates moderate cross-modal alignment, falling within the typical range of 0.15--0.80 reported in (satellite) imagery and vision-language alignment literature \citep{dhakal2024sat2cap,radford2021learning}. We interpret this result as suggestive evidence for partial alignment, not as decisive evidence for the Platonic Representation Hypothesis.

\FloatBarrier

\paragraph{Regional Representational Convergence.}

To explore the Platonic Representation Hypothesis in a geographically informed manner, we investigate whether representational alignment between vision/language modalities varies by region, potentially reflecting differences in socio-economic documentation or infrastructure visibility across Africa. 

\begin{figure}[!htbp]
  \centering
  \includegraphics[width=0.65\linewidth]{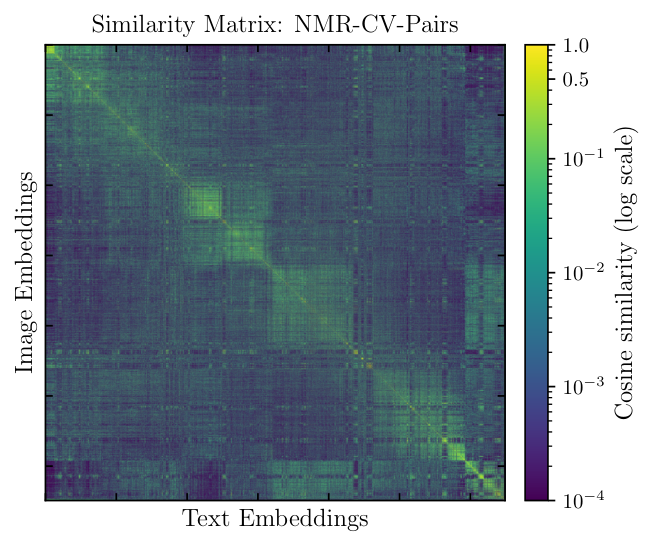}
  \caption{
    Latent representation similarity matrix of matching embeddings from NMR and CV, based on cosine similarity after alignment through canonical correlation. Embeddings are sorted by latitude to allow regional interpretability. See Figure \ref{fig:CosineSimMatrix_ASA_CV} for ASA/CV comparison. 
  }
  \label{fig:CosineSimMatrix}
\end{figure}

To this end, we compute a similarity matrix between NMR- and CV-based embeddings using cosine similarity after alignment through canonical correlation analysis (CCA) \citep{weenink2003canonical}, with rows/columns sorted by latitude for regional interpretability (Figure \ref{fig:CosineSimMatrix}). The resulting matrix reveals localized clusters of higher similarity, suggesting that embeddings from geographically proximate regions along similar latitudes exhibit stronger alignment. This spatial coherence indicates that geographically proximate areas exhibit systematically stronger cross-modal alignment.

\paragraph{Additional Robustness Checks.} Appendix I provides details on additional robustness tests. For example, it examines potential training/test leakage. There, in Figure \ref{fig:r2_years}, we analyze whether temporal patterns in model performance might indicate leakage into future periods. As shown in Figure~\ref{fig:r2_years}, $R^2$ scores across years for exemplary models (\texttt{ASA-CleanedTraces-OAI3Small} and \texttt{NMR-Llama4 Maverick-OAI3Small}) exhibit no clear temporal trend, indicating consistent performance over time and minimal evidence of direct data leakage.

\FloatBarrier

\clearpage 

\section{Discussion \& Conclusion}\label{s:Conclusion}

Overall, our findings suggest that socio-economic indicators, such as household wealth, leave recoverable signals in both visual and textual modalities: fused models outperform single-modality baselines by 21.77\% in $R^2$ across random and country-holdout evaluations. Analysis within our framework indicates moderate alignment between vision and text embeddings, which exhibit a median cosine similarity of around 0.60 after alignment through canonical correlation analysis---a pattern broadly consistent with, but not decisive evidence for, the Platonic Representation Hypothesis \citep{huh2024position}. Evidence for agent-induced novelty is limited. By releasing a large-scale multimodal DHS dataset, we enable further research in AI-augmented poverty mapping, with implications for equitable global development monitoring and policy interventions.

From a practical standpoint, the NMR approach offers significant scalability advantages over the ASA approach. The NMR pipeline requires only a single LLM inference per location, whereas the ASA approach requires a full web search and text processing pipeline. This makes NMR more cost-effective and suitable for large-scale poverty mapping applications. Our results show that NMR alone achieves 95\% of the performance of the combined NMR+CV approach, making it an excellent choice for resource-constrained deployments. 

Limitations of our study include reliance on DHS clusters, which may introduce sampling biases, as remote areas are excluded from sampling \citep{kakooei2024increasing}. The search agent may also retrieve post-treatment information from the web, potentially biasing causal analysis. Also, the computational demands of agent-driven pipelines at continental scales limit scalability for global applications. 

Future work could integrate causal inference into multimodal AI, addressing text-specific challenges \citep{grimmer2022text,daoud-etal-2022-conceptualizing,pieuchon2024can}---such as improved detection and control of post-treatment bias in agent representations---while further testing the Platonic Representation and Agent-Induced Novelty Hypotheses in open-ended contexts.

\begin{credits}
\subsubsection{\ackname} The authors thank members of the AI \& Global Development Lab for helpful feedback. 

\subsubsection{\discintname}
The authors have no competing interests to declare that are relevant to the content of this article.
\end{credits}

% bibliography print 
%\printbibliography
\bibliographystyle{splncs04}
\bibliography{mybib}

\begin{thebibliography}{10}
\providecommand{\url}[1]{\texttt{#1}}
\providecommand{\urlprefix}{URL }
\providecommand{\doi}[1]{https://doi.org/#1}

\bibitem{aiken2023fairness}
Aiken, E., Rolf, E., Blumenstock, J.: Fairness and representation in
  satellite-based poverty maps: Evidence of urban-rural disparities and their
  impacts on downstream policy. arXiv preprint arXiv:2305.01783  (2023),
  \url{https://arxiv.org/abs/2305.01783}, {IJCAI 2023 AI for Social Good Track}

\bibitem{arzaghi2024understandingintrinsicsocioeconomicbiases}
Arzaghi, M., Carichon, F., Farnadi, G.: Understanding intrinsic socioeconomic
  biases in large language models (2024),
  \url{https://arxiv.org/abs/2405.18662}

\bibitem{pieuchon2024can}
{Audinet de Pieuchon}, N., Daoud, A., Jerzak, C., Johansson, M., Johansson, R.:
  Can large language models (or humans) disentangle text? In: Proceedings of
  the Sixth Workshop on Natural Language Processing and Computational Social
  Science (NLP+CSS 2024). pp. 57--67. Association for Computational
  Linguistics, Mexico City, Mexico (2024). \doi{10.18653/v1/2024.nlpcss-1.5}

\bibitem{depieuchon2025benchmarkingdebiasingmethodsllmbased}
{Audinet de Pieuchon}, N., Daoud, A., Jerzak, C.T., Johansson, M., Johansson,
  R.: Benchmarking debiasing methods for llm-based parameter estimates (2025),
  \url{https://arxiv.org/abs/2506.09627}, proceedings of the 2025 Conference on
  Empirical Methods in Natural Language Processing (EMNLP)

\bibitem{babenko2017povertymappingusingconvolutional}
Babenko, B., Hersh, J., Newhouse, D., Ramakrishnan, A., Swartz, T.: Poverty
  mapping using convolutional neural networks trained on high and medium
  resolution satellite images, with an application in mexico (2017),
  \url{https://arxiv.org/abs/1711.06323}, presented at the NIPS 2017 Workshop
  on Machine Learning for the Developing World

\bibitem{burkeUsingSatelliteImagery2021}
Burke, M., Driscoll, A., Lobell, D.B., Ermon, S.: Using satellite imagery to
  understand and promote sustainable development. Science  \textbf{371}(6535),
  eabe8628 (2021). \doi{10.1126/science.abe8628}

\bibitem{daoudStatisticalModelingThree2023}
Daoud, A., Dubhashi, D.: Statistical {{Modeling}}: {{The Three Cultures}}.
  Harvard Data Science Review  \textbf{5}(1) (2023).
  \doi{10.1162/99608f92.89f6fe66}, winter 2023

\bibitem{daoud-etal-2022-conceptualizing}
Daoud, A., Jerzak, C., Johansson, R.: Conceptualizing treatment leakage in
  text-based causal inference. In: Carpuat, M., de~Marneffe, M.C., Meza~Ruiz,
  I.V. (eds.) Proceedings of the 2022 Conference of the North American Chapter
  of the Association for Computational Linguistics: Human Language
  Technologies. pp. 5638--5645. Association for Computational Linguistics,
  Seattle, United States (Jul 2022). \doi{10.18653/v1/2022.naacl-main.413},
  \url{https://aclanthology.org/2022.naacl-main.413/}

\bibitem{daoud2019international}
Daoud, A., Reinsberg, B., Kentikelenis, A.E., Stubbs, T.H., King, L.P.: The
  international monetary fund’s interventions in food and agriculture: An
  analysis of loans and conditions. Food Policy  \textbf{83},  204--218 (2019).
  \doi{10.1016/j.foodpol.2019.01.005}

\bibitem{dhakal2024sat2cap}
Dhakal, A., Ahmad, A., Khanal, S., Sastry, S., Kerner, H., Jacobs, N.:
  {Sat2Cap}: Mapping fine-grained textual descriptions from satellite images.
  In: Proceedings of the IEEE/CVF Conference on Computer Vision and Pattern
  Recognition Workshops. pp. 533--542 (2024)

\bibitem{elvidge2009global}
Elvidge, C.D., Sutton, P.C., Tuttle, B.T., Ghosh, T., Baugh, K.E.: Global urban
  mapping based on nighttime lights. In: Gamba, P., Herold, M. (eds.) Global
  Mapping of Human Settlement: Experiences, Datasets, and Prospects, chap.~6,
  pp. 129--144. CRC Press, Boca Raton, FL (2009)

\bibitem{gema2025inverse}
Gema, A.P., H{\"a}gele, A., Chen, R., Arditi, A., Goldman-Wetzler, J.,
  Fraser-Taliente, K., Sleight, H., Petrini, L., Michael, J., Alex, B.,
  Minervini, P., Chen, Y., Benton, J., Perez, E.: Inverse scaling in test-time
  compute. Transactions on Machine Learning Research  (2025),
  \url{https://openreview.net/forum?id=NXgyHW1c7M}, published December 2025;
  arXiv:2507.14417

\bibitem{grimmer2022text}
Grimmer, J., Roberts, M.E., Stewart, B.M.: Text as data: A new framework for
  machine learning and the social sciences. Princeton University Press (2022)

\bibitem{han2024geosee}
Han, S., Ahn, D., Lee, S., Song, M., Park, S., Park, S., Kim, J., Cha, M.:
  Geosee: Regional socio-economic estimation with a large language model. arXiv
  preprint arXiv:2406.09799  (2024)

\bibitem{huh2024position}
Huh, M., Cheung, B., Wang, T., Isola, P.: Position: The {Platonic
  Representation Hypothesis}. In: Salakhutdinov, R., Kolter, Z., Heller, K.,
  Weller, A., Oliver, N., Scarlett, J., Berkenkamp, F. (eds.) Proceedings of
  the 41st International Conference on Machine Learning. Proceedings of Machine
  Learning Research, vol.~235, pp. 20617--20642. PMLR (2024),
  \url{https://proceedings.mlr.press/v235/huh24a.html}

\bibitem{jean2016combining}
Jean, N., Burke, M., Xie, M., Alampay~Davis, W.M., Lobell, D.B., Ermon, S.:
  Combining satellite imagery and machine learning to predict poverty. Science
  \textbf{353}(6301),  790--794 (2016). \doi{10.1126/science.aaf7894}

\bibitem{jerzak2023integrating}
Jerzak, C.T., Johansson, F., Daoud, A.: Integrating earth observation data into
  causal inference: challenges and opportunities. arXiv preprint
  arXiv:2301.12985  (2023)

\bibitem{JungLastMile}
Jung, W., Sinha, A., Kim, A., Shah, V., Lu, Y., Lee, L., Ammari, T.: The last
  mile in remote sensing poverty prediction. ACM J. Comput. Sustain. Soc.
  \textbf{3}(3),  1--54 (Jun 2025). \doi{10.1145/3724422}, article 16

\bibitem{kakooei2024mappingafricasettlementshigh}
Kakooei, M., Bailie, J., Pettersson, M.B., Söderberg, A., Becevic, A., Daoud,
  A.: A high resolution urban and rural settlement map of africa using deep
  learning and satellite imagery. Scientific Reports  \textbf{16}, 637 (2026).
  \doi{10.1038/s41598-025-34295-7}

\bibitem{kakooei2024increasing}
Kakooei, M., Daoud, A.: Increasing the confidence of predictive uncertainty:
  earth observations and deep learning for poverty estimation. IEEE
  Transactions on Geoscience and Remote Sensing  \textbf{62},  1--13 (2024).
  \doi{10.1109/TGRS.2024.3392605}, article 4704613

\bibitem{kakooei2024analyzingpovertyintraannualtimeseries}
Kakooei, M., Solska, K., Daoud, A.: Analyzing poverty through intra-annual
  time-series: A wavelet transform approach (2024),
  \url{https://arxiv.org/abs/2411.02855}

\bibitem{lamichhane2025exploring}
Lamichhane, B.R., Isnan, M., Horanont, T.: Exploring machine learning trends in
  poverty mapping: A review and meta-analysis. Science of Remote Sensing
  \textbf{11}, 100200 (2025). \doi{10.1016/j.srs.2025.100200}

\bibitem{dhs2013demographic}
{MEASURE DHS}: Demographic and health surveys. Report, {MEASURE DHS},
  Calverton, MD (2013), \url{http://pdf.usaid.gov/pdf_docs/pnaeb738.pdf}

\bibitem{miehling2025agentic}
Miehling, E., Ramamurthy, K.N., Varshney, K.R., Riemer, M., Bouneffouf, D.,
  Richards, J.T., Dhurandhar, A., Daly, E.M., Hind, M., Sattigeri, P., et~al.:
  Agentic {AI} needs a systems theory. arXiv preprint arXiv:2503.00237  (2025)

\bibitem{o2023seeing}
O'Brien, J.D.: Seeing What We Can't: Evaluating implicit biases in deep
  learning satellite imagery models trained for poverty prediction. Honors
  thesis, William \& Mary, Williamsburg, VA (2023),
  \url{https://scholarworks.wm.edu/entities/publication/ab00bf9c-f7a6-49bd-8709-f3806768587b}

\bibitem{pettersson2023time}
Pettersson, M.B., Kakooei, M., Ortheden, J., Johansson, F.D., Daoud, A.: Time
  series of satellite imagery improve deep learning estimates of
  neighborhood-level poverty in africa. In: Proceedings of the Thirty-Second
  International Joint Conference on Artificial Intelligence. pp. 6165--6173.
  International Joint Conferences on Artificial Intelligence Organization
  (2023). \doi{10.24963/ijcai.2023/684}, {AI for Good}

\bibitem{radford2021learning}
Radford, A., Kim, J.W., Hallacy, C., Ramesh, A., Goh, G., Agarwal, S., Sastry,
  G., Askell, A., Mishkin, P., Clark, J., Krueger, G., Sutskever, I.: Learning
  transferable visual models from natural language supervision. In: Meila, M.,
  Zhang, T. (eds.) Proceedings of the 38th International Conference on Machine
  Learning. Proceedings of Machine Learning Research, vol.~139, pp. 8748--8763.
  PMLR (2021), \url{https://proceedings.mlr.press/v139/radford21a.html}

\bibitem{sakamoto2025scopingreviewearthobservation}
Sakamoto, K., Jerzak, C.T., Daoud, A.: A scoping review of earth observation
  and machine learning for causal inference: Implications for the geography of
  poverty. In: Hall, O., Wahab, I. (eds.) Geography of Poverty. Edward Elgar
  Publishing, Cheltenham, UK (2025), \url{https://arxiv.org/abs/2406.02584}

\bibitem{sarmadi2025leveragingchatgptsmultimodalvision}
Sarmadi, H., Hall, O., Rögnvaldsson, T., Ohlsson, M.: Leveraging chatgpt's
  multimodal vision capabilities to rank satellite images by poverty level:
  Advancing tools for social science research (2025),
  \url{https://arxiv.org/abs/2501.14546}

\bibitem{stewart2022torchgeo}
Stewart, A.J., Robinson, C., Corley, I.A., Ortiz, A., {Lavista Ferres}, J.M.,
  Banerjee, A.: {TorchGeo}: Deep learning with geospatial data. In: Proceedings
  of the 30th International Conference on Advances in Geographic Information
  Systems. pp. 1--12. Association for Computing Machinery, New York, NY, USA
  (2022). \doi{10.1145/3557915.3560953}

\bibitem{wang2024agent}
Wang, J., Duan, Z.: Agent {AI} with {LangGraph}: A modular framework for
  enhancing machine translation using large language models. arXiv preprint
  arXiv:2412.03801  (2024), \url{https://arxiv.org/abs/2412.03801}

\bibitem{weenink2003canonical}
Weenink, D.J.M.: Canonical correlation analysis. In: Proceedings of the
  Institute of Phonetic Sciences of the University of Amsterdam. vol.~25, pp.
  81--99 (2003)

\bibitem{yeh2020using}
Yeh, C., Perez, A., Driscoll, A., Azzari, G., Tang, Z., Lobell, D., Ermon, S.,
  Burke, M.: Using publicly available satellite imagery and deep learning to
  understand economic well-being in africa. Nature Communications
  \textbf{11}(1), 2583 (2020). \doi{10.1038/s41467-020-16185-w}

\bibitem{zhou2024synergizingllmagentsknowledge}
Zhou, Z., Fan, J., Liu, Y., Xu, F., Jin, D., Li, Y.: Synergizing llm agents and
  knowledge graph for socioeconomic prediction in lbsn (2024),
  \url{https://arxiv.org/abs/2411.00028}

\bibitem{pmlr-v275-zhu25a}
Zhu, F.W., Jerzak, C.T., Daoud, A.: Optimizing multi-scale representations to
  detect effect heterogeneity using earth observation and computer vision:
  Applications to two anti-poverty rcts. In: Huang, B., Drton, M. (eds.)
  Proceedings of the Fourth Conference on Causal Learning and Reasoning.
  Proceedings of Machine Learning Research, vol.~275, pp. 894--919. PMLR
  (07--09 May 2025), \url{https://proceedings.mlr.press/v275/zhu25a.html}

\end{thebibliography}
%\newpage \clearpage 

\renewcommand\thefigure{A.I.\arabic{figure}} \setcounter{figure}{0}
\renewcommand\thetable{A.I.\arabic{table}} \setcounter{table}{0}

\section{Appendix I. Additional Empirical Results}

\begin{table}[htb]
  \centering
  \caption{Test Split Performance ($R^2$ and RMSE) Across Evaluation Strategies, Grouped by Modality and Embedding Model, Sorted by Random Split $R^2$}
  \label{tab:test_split_sorted_by_modality}
  \begin{tabular}{l l l c c c c c c}
    \hline
    \textbf{Procedure} & \textbf{Source} & \textbf{Embedding} &
    \textbf{R²} & \textbf{RMSE} & \textbf{R²} & \textbf{RMSE} & \textbf{R²} & \textbf{RMSE} \\
    & & & \multicolumn{2}{c}{\textbf{Random Split}} & \multicolumn{2}{c}{\textbf{OOC Split}} & \multicolumn{2}{c}{\textbf{OOT Split}} \\
    \hline
    NMR+ASA+CV & Llama+Traces & OAI+ViT & 0.772 & 9.045 & 0.529 & 12.169 & 0.694 & 10.165 \\
    NMR+CV & Llama4Maverick & OAI+ViT & 0.765 & 9.196 & 0.527 & 12.132 & 0.700 & 10.075 \\
    NMR+ASA+CV & Llama+Traces & MPNet+ViT & 0.754 & 9.412 & 0.504 & 12.288 & 0.701 & 9.947 \\
    NMR+CV & Llama4Maverick & MPNet+ViT & 0.747 & 9.533 & 0.536 & 12.085 & 0.698 & 9.998 \\
    ASA+CV & CleanedTraces & OAI+ViT & 0.740 & 9.662 & 0.463 & 12.893 & 0.659 & 10.735 \\
    ASA+CV & CleanedTraces & MPNet+ViT & 0.698 & 10.423 & 0.487 & 12.808 & 0.628 & 11.093 \\
    NMR+ASA & Llama+Traces & OAI3Small & 0.697 & 10.460 & 0.371 & 14.062 & 0.610 & 11.516 \\
    NMR & Llama4Maverick & OAI3Small & 0.668 & 10.945 & 0.440 & 13.285 & 0.609 & 11.527 \\
    NMR+ASA & Llama+Traces & MPNet-FT & 0.667 & 10.961 & 0.386 & 13.871 & 0.605 & 11.478 \\
    NMR & Llama4Maverick & MPNet-FT & 0.641 & 11.456 & 0.418 & 13.700 & 0.558 & 12.135 \\
    CV & EOData & ViT16Small & 0.634 & 11.469 & 0.446 & 13.194 & 0.569 & 12.080 \\
    NMR & GPT4.1Nano & OAI3Small & 0.622 & 11.680 & 0.362 & 14.192 & 0.557 & 12.288 \\
    NMR & DeepSeekR1 & MPNet-FT & 0.607 & 11.984 & 0.417 & 13.792 & 0.523 & 12.611 \\
    ASA & CleanedTraces & OAI3Small & 0.606 & 11.921 & 0.301 & 14.739 & 0.504 & 12.973 \\
    NMR & GPT4.1Nano & MPNet-FT & 0.591 & 12.231 & 0.351 & 14.543 & 0.504 & 12.886 \\
    ASA & Top10Search & OAI3Small & 0.588 & 12.197 & 0.262 & 15.148 & 0.483 & 13.248 \\
    ASA & Wikipedia & OAI3Small & 0.565 & 12.529 & 0.225 & 15.618 & 0.455 & 13.605 \\
    ASA & JustificationOnly & OAI3Small & 0.557 & 12.645 & 0.279 & 15.057 & 0.463 & 13.514 \\
    ASA & JustificationPrediction & OAI3Small & 0.544 & 12.832 & 0.287 & 14.969 & 0.458 & 13.585 \\
    NMR & Grok3Mini & MPNet-FT & 0.503 & 13.482 & 0.254 & 15.562 & 0.394 & 14.261 \\
    NMR & Grok3Mini & OAI3Small & 0.493 & 13.532 & 0.269 & 15.111 & 0.438 & 13.819 \\
    ASA & CleanedTraces & MPNet-FT & 0.478 & 13.722 & 0.241 & 15.717 & 0.382 & 14.344 \\
    ASA & Wikipedia & MPNet-FT & 0.477 & 13.735 & 0.138 & 16.677 & 0.377 & 14.402 \\
    ASA & Top10Search & MPNet-FT & 0.475 & 13.764 & 0.229 & 15.865 & 0.381 & 14.345 \\
    ASA & JustificationPrediction & MPNet-FT & 0.450 & 14.092 & 0.230 & 15.826 & 0.374 & 14.465 \\
    ASA & JustificationOnly & MPNet-FT & 0.447 & 14.131 & 0.220 & 15.910 & 0.371 & 14.504 \\
    NMR & DeepSeekR1 & OAI3Small & 0.008 & 18.918 & -0.143 & 19.055 & -0.035 & 18.911 \\
    \hline
  \end{tabular}
\end{table}

% \clearpage \newpage
\begin{table*}[htb]
\centering
\scriptsize
\caption{95\% confidence intervals for $R^2$ and RMSE across evaluation strategies, grouped by modality and embedding model. Intervals are quantile-based, from bootstrapping through entire train/test split and estimation procedures. }
\label{tab:ci95_bounds_summary}
\begingroup
\setlength{\tabcolsep}{2pt}
\renewcommand{\arraystretch}{0.96}
\begin{adjustwidth}{-2cm}{-2cm}
\centering
\resizebox{\linewidth}{!}{%
\begin{tabular}{l l l c c c c c c}
\hline
\textbf{Procedure} & \textbf{Source} & \textbf{Embedding} &
\textbf{R²} & \textbf{RMSE} & \textbf{R²} & \textbf{RMSE} & \textbf{R²} & \textbf{RMSE} \\
& & & \multicolumn{2}{c}{\textbf{Random Split}} & \multicolumn{2}{c}{\textbf{OOC Split}} & \multicolumn{2}{c}{\textbf{OOT Split}} \\
\hline
NMR+ASA+CV & Llama+Traces & OAI+ViT & [0.759; 0.781] & [8.780; 9.212] & [0.232; 0.737] & [9.517; 16.411] & [0.594; 0.803] & [9.323; 11.211] \\
NMR+CV & Llama4Maverick & OAI+ViT & [0.755; 0.777] & [9.011; 9.270] & [0.344; 0.722] & [10.561; 14.512] & [0.608; 0.774] & [9.421; 11.070] \\
NMR+ASA+CV & Llama+Traces & MPNet+ViT & [0.748; 0.760] & [9.308; 9.557] & [0.171; 0.629] & [10.537; 13.762] & [0.643; 0.765] & [9.695; 10.309] \\
NMR+CV & Llama4Maverick & MPNet+ViT & [0.742; 0.753] & [9.430; 9.662] & [0.386; 0.624] & [10.222; 13.904] & [0.630; 0.765] & [9.606; 10.488] \\
ASA+CV & CleanedTraces & OAI+ViT & [0.729; 0.761] & [9.521; 9.988] & [0.422; 0.541] & [11.433; 13.914] & [0.568; 0.689] & [10.021; 11.344] \\
ASA+CV & CleanedTraces & MPNet+ViT & [0.688; 0.705] & [10.285; 10.605] & [0.448; 0.538] & [11.233; 14.945] & [0.537; 0.706] & [10.693; 11.737] \\
NMR+ASA & Llama+Traces & OAI3Small & [0.696; 0.698] & [10.442; 10.479] & [0.344; 0.405] & [13.722; 14.319] & [0.662; 0.684] & [11.436; 11.599] \\
NMR & Llama4Maverick & OAI3Small & [0.667; 0.669] & [10.926; 10.964] & [0.423; 0.457] & [13.008; 13.561] & [0.596; 0.622] & [11.447; 11.607] \\
NMR+ASA & Llama+Traces & MPNet-FT & [0.657; 0.675] & [10.800; 11.106] & [0.084; 0.606] & [12.638; 15.234] & [0.521; 0.689] & [11.133; 11.982] \\
NMR & Llama4Maverick & MPNet-FT & [0.636; 0.646] & [11.367; 11.545] & [0.351; 0.486] & [12.870; 14.530] & [0.485; 0.630] & [11.805; 12.464] \\
CV & EOData & ViT16Small & [0.628; 0.636] & [11.354; 11.511] & [0.433; 0.464] & [12.959; 14.259] & [0.524; 0.601] & [11.892; 12.122] \\
NMR & GPT4.1Nano & OAI3Small & [0.621; 0.623] & [11.660; 11.700] & [0.344; 0.381] & [13.893; 14.492] & [0.543; 0.570] & [12.204; 12.371] \\
NMR & DeepSeekR1 & MPNet-FT & [0.600; 0.615] & [11.884; 12.083] & [0.375; 0.459] & [12.386; 15.198] & [0.451; 0.596] & [12.249; 12.974] \\
ASA & CleanedTraces & OAI3Small & [0.605; 0.608] & [11.901; 11.940] & [0.271; 0.332] & [14.484; 14.994] & [0.486; 0.522] & [12.873; 13.074] \\
NMR & GPT4.1Nano & MPNet-FT & [0.586; 0.595] & [12.151; 12.311] & [0.287; 0.416] & [12.958; 16.128] & [0.440; 0.568] & [12.565; 10.533] \\
ASA & Top10Search & OAI3Small & [0.587; 0.589] & [12.179; 12.216] & [0.231; 0.292] & [14.917; 15.380] & [0.465; 0.501] & [13.151; 13.344] \\
ASA & Wikipedia & OAI3Small & [0.564; 0.567] & [12.509; 12.548] & [0.201; 0.249] & [15.328; 15.907] & [0.436; 0.474] & [13.501; 13.708] \\
ASA & JustificationOnly & OAI3Small & [0.556; 0.559] & [12.624; 12.666] & [0.257; 0.300] & [14.819; 15.295] & [0.445; 0.480] & [13.427; 13.601] \\
ASA & JustificationPrediction & OAI3Small & [0.542; 0.545] & [12.811; 12.853] & [0.265; 0.309] & [14.727; 15.211] & [0.441; 0.475] & [13.505; 13.665] \\
NMR & Grok3Mini & MPNet-FT & [0.497; 0.508] & [13.388; 13.577] & [0.172; 0.335] & [14.124; 17.000] & [0.327; 0.462] & [13.713; 14.809] \\
NMR & Grok3Mini & OAI3Small & [0.491; 0.494] & [13.509; 13.554] & [0.242; 0.297] & [14.833; 15.389] & [0.418; 0.458] & [13.644; 13.994] \\
ASA & CleanedTraces & MPNet-FT & [0.472; 0.485] & [13.605; 13.839] & [0.196; 0.287] & [14.327; 17.106] & [0.279; 0.485] & [13.926; 14.763] \\
ASA & Wikipedia & MPNet-FT & [0.473; 0.481] & [13.701; 13.768] & [0.035; 0.241] & [15.583; 17.772] & [0.275; 0.479] & [13.994; 14.810] \\
ASA & Top10Search & MPNet-FT & [0.470; 0.480] & [13.657; 13.871] & [0.183; 0.275] & [14.266; 17.464] & [0.275; 0.487] & [13.887; 14.804] \\
ASA & JustificationPrediction & MPNet-FT & [0.442; 0.458] & [14.005; 14.178] & [0.161; 0.300] & [14.300; 17.351] & [0.287; 0.461] & [14.166; 14.764] \\
ASA & JustificationOnly & MPNet-FT & [0.437; 0.456] & [14.037; 14.224] & [0.147; 0.293] & [14.584; 17.235] & [0.285; 0.235] & [14.193; 14.815] \\
NMR & DeepSeekR1 & OAI3Small & [0.008; 0.009] & [18.898; 18.939] & [-0.174; -0.113] & [18.608; 19.502] & [-0.041; -0.029] & [18.668; 19.154] \\
\hline
\end{tabular}
}%
\end{adjustwidth}
\endgroup
\end{table*}

\clearpage \newpage

\begin{figure}[htb]
  \centering
  \includegraphics[width=0.9\linewidth]{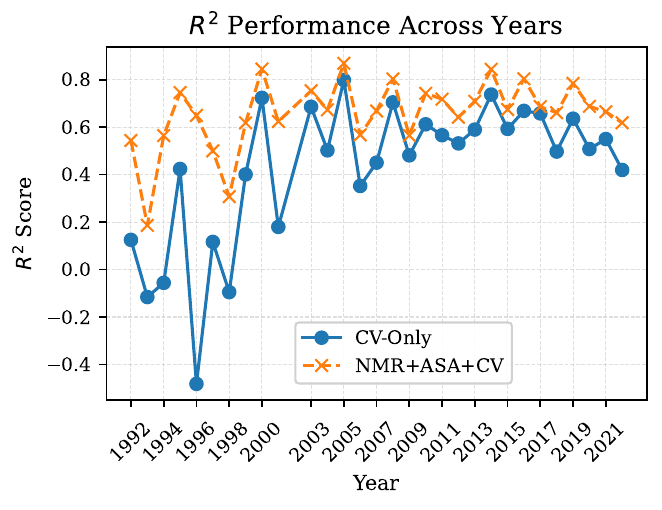}
  \caption{
    $R^2$ scores across years for the best system combining NMR+ASA+CV benchmarking against CV-only.
    }
  \label{fig:r2_years_cv_comb}
\end{figure}

\paragraph{Data Leakage Considerations.}

It is possible that IWI information from the DHS data appears within the agent traces or the neural weights. This does not cause train/test leakage unless the data postdates the model's training snapshot. Moreover, the DHS data are not publicly available, and to our knowledge, are not directly used in training corpora. Moreover, the MPNet model used in many of our analyses was trained in 2020, predating much of the DHS data examined here.

\begin{figure}[htb]
  \centering
  \includegraphics[width=0.9\linewidth]{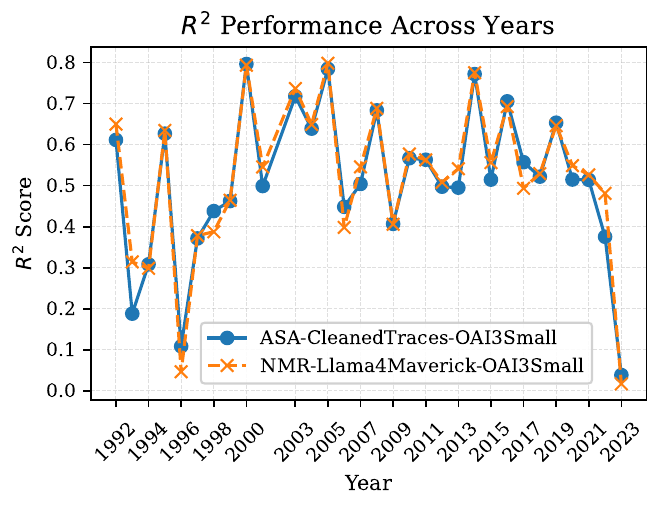}
  \caption{
    $R^2$ scores across years for two exemplary model variants:
    \texttt{ASA-CleanedTraces-OAI3Small} and \texttt{NMR-Llama4Maverick-OAI3Small}. No clear temporal trend is observed (as we might have expected under direct data leakage, assuming 
    that more recent DHS surveys are systematically more likely or less likely to be included in the LLM training corpora.
  }
  \label{fig:r2_years}
\end{figure}

While we cannot definitively determine whether such information was used in model training, we did perform a string search of the agent traces for the terms ``IWI,'' ``International Wealth Index,'' and ``DHS.'' We found that 10.4\% of agent traces contained these query terms, which could be associated with documentary evidence from both pre- and post-focal year. We reran the ASA results without these observations and found similar results as with (\texttt{ASA-CleanedTraces-OAI3Small}: $R^2_{\textsc{Random}}=0.585$; RMSE$_{\textsc{Random}}=12.229$); see also Figure \ref{fig:r2_years} and Figure \ref{fig:residuals_years_cv_comb}, which shows absolute improvements from multimodal data over time relative the image-only baseline. 

\begin{figure}[htb!]
  \centering
  \includegraphics[width=0.65\linewidth]{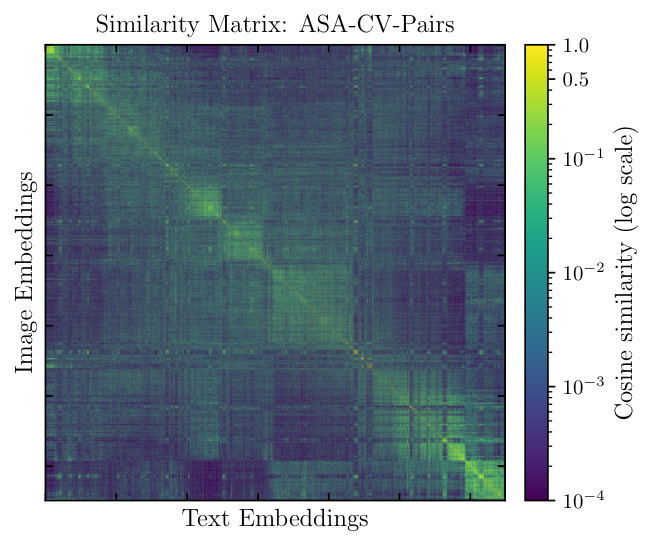}
  \caption{
    Latent representation similarity matrix of matching embeddings from ASA and CV, based on cosine similarity after alignment through canonical correlation analysis. Embeddings are sorted by latitude to allow regional interpretability across pairs.
  }
  \label{fig:CosineSimMatrix_ASA_CV}
\end{figure}

%\clearpage 
%\newpage \newpage

\renewcommand\thefigure{A.II.\arabic{figure}} \setcounter{figure}{0}
\renewcommand\thetable{A.II.\arabic{table}} \setcounter{table}{0}

\section{Appendix II. Modeling Details} 
\paragraph{Computing Infrastructure.}
All experiments presented were generated on the following hardware/software setup:
\begin{itemize}
    \item 4 Tesla V100-PCIE-32GB; total VRAM: 128GB. (CUDA Version: 12.6).
    \item 80 Intel(R) Xeon(R) Gold 6148 CPUs @ 2.40GHz; total RAM: 754GB.
    \item Operating System: Ubuntu 24.04.2 LTS.
    \item Kernel: Linux 6.8.0-57-generic.
    \item Python Version: 3.10.15.
\end{itemize}

\paragraph{Pretrained Model Details.} The used LLMs are described in Table\ref{tab:llm_details} and information on the embedding models chosen is to be found in Table\ref{tab:emb_details}

\begin{table}[htb]
  \centering
  \small
  \caption{Details of the LLMs used in our experiments. Model size is shown if available.}
  \label{tab:llm_details}
  \begin{tabular}{l c c c}
    \hline
    \textbf{Model} & \textbf{Developer} & \textbf{Size} & \textbf{API} \\
    \hline
    GPT-4.1 Nano & OpenAI & --- & OpenAI \\
    Llama 4 Maverick & Meta & 405B (A17B) & Groq \\
    DeepSeek R1 (0528) & DeepSeek & 671B (A37B) & DeepSeek \\
    Grok 3 Mini & xAI & --- & xAI \\
    \hline
  \end{tabular}
\end{table}

\begin{table}[htb]
  \centering
  \small
  \caption{Details of embedding models used.}
  \label{tab:emb_details}
  \begin{tabular}{l c c}
    \hline
    \textbf{Model} & \textbf{Context Len. (Tokens)} & \textbf{Emb. Dim} \\
    \hline
    OpenAI (text-embed-3-small) & 8192 & 1536 \\
    MPNet (all-mpnet-base-v2) & 384 & 768 \\
    \hline
  \end{tabular}
\end{table}

To encode the satellite imagery, a pretrained Vision Transformer model, 
\begin{itemize}
    \item[] \texttt{ViTSmall16$\_$Weights.LANDSAT$\_$ETM$\_$SR$\_$MOCO},
\end{itemize}
is deployed and pre-trained on Landsat imagery \citep{stewart2022torchgeo}. Satellite image inputs were 5-channel (RGB plus shortwave and longwave infrared) and 224$\times$224 resolution. We mask clouds and cloud shadows using Landsat quality-assurance bands and form median temporal composites centered on each survey year. Ridge regression was employed for supervised prediction from visual, textual, and fused embeddings, with ridge penalty $\alpha=1.0$. The only model fine-tuned beyond ridge regression was the MPNet embedding model (all-mpnet-base-v2), which was trained using 5-fold cross-validation. All other embeddings (e.g., OpenAI text-embed-3-small) were used in frozen mode (no fine-tuning). Joint representations were constructed by directly concatenating embeddings, and the differing dimensions (e.g., 768 for MPNet, 1536 for OpenAI embeddings) were handled without requiring a projection. The AI Search Agent was built using LangGraph, with a maximum recursion depth of 20. All experiments used 80/20 train/test partitions, with country-level units for OOC splits and year-level units for OOT splits. 

\end{document}